\documentclass[conference]{IEEEtran}
\IEEEoverridecommandlockouts
\usepackage[square,numbers]{natbib}
\usepackage{float}
\usepackage[pdftex]{graphicx}
	\DeclareGraphicsExtensions{.pdf,.jpeg,.png,.jpg}
    \graphicspath{{./images/}}
\usepackage{amsmath}
\usepackage{algorithmic}
\usepackage{comment}
\usepackage{booktabs}
\usepackage{subcaption}
\usepackage{array}
\usepackage{tikz}
\usetikzlibrary{arrows}
\usetikzlibrary{shapes}
\usetikzlibrary{shapes.geometric}
\usetikzlibrary{fit, positioning}
\usepackage{multirow}

\usepackage{url}
	
\title{Image Enhancement and Object Recognition for Night Vision Surveillance}
\author{\IEEEauthorblockN{Aashish Bhandari\textsuperscript{*}, Aayush Kafle\textsuperscript{*}, Pranjal Dhakal\textsuperscript{*}, Prateek Raj Joshi\textsuperscript{*} and Dinesh Baniya Kshatri}
\IEEEauthorblockA{\textit{Department of Electronics and Computer Engineering}\\ \textit{IoE, Tribhuvan University}\\Lalitpur, Nepal} \thanks{* They have equal contribution in this work}}
\begin{document}
\maketitle
\begin{abstract}
\par
Object recognition is a critical part of any surveillance system. It is the matter of utmost concern to identify intruders and foreign objects in the area where surveillance is done. The performance of surveillance system using the traditional camera in daylight is vastly superior as compared to night. The main problem for surveillance during the night is the objects captured by traditional cameras have low contrast against the background because of the absence of ambient light in the visible spectrum. Due to that reason, the image is taken in low light condition using an Infrared Camera and the image is enhanced to obtain an image with higher contrast using different enhancing algorithms based on the spatial domain. The enhanced image is then sent to the classification process. The classification is done by using convolutional neural network followed by a fully connected layer of neurons. The accuracy of classification after implementing different enhancement algorithms is compared in this paper.
 \end{abstract}




\section{Introduction}\label{section:introduction}
\par
A security surveillance system is incomplete without automatic classification of objects that are in the area. The development and discoveries in image processing have been a great help in recognizing objects for surveillance. However, recognizing objects in the night still remains a challenge. 
\par
The inability of human vision to see in dark has not only limited our work efficiency, it also had increased crime and offense in night time. There are systems that detect the objects and their movement in the night time but they suffer from inaccurate prediction when recognizing the objects and people.
\par
Numerous research has been carried out in object recognition for fairly illuminated images achieving an acceptable level of accuracy. Yet, the research and findings for automated classification of low light and no light images have been few and far.
\par
In a technique known as "Active Illumination",\cite{grossman_2007} the dark environment is illuminated using near-infrared(NIR) of wavelength 700-1000 nm which is just below the visible spectrum\cite{ISO20473}. Modern Charge Coupled Devices are capable sensitive to this wavelength and therefore capable of capturing the images lit by NIR LED. The resultant images still lack the contrast and brightness being monochromatic in nature. This is the prime reason for using different image enhancement techniques based on spatial domain. 
\par
Variety of histogram equalization methods can be used to perform contrast stretching in dimly-lit image which improves the visual appeal and also increases the details of objects in images. Global contrast equalization technique can be implemented for enhancing video quality of night-vision cameras. Metrics like Peak Signal to Noise Ratio(PSNR) is used to quantitatively justify the enhancement in image\cite{ramos_2015}. Different derivatives of histogram equalization are being used to enhance the image by preserving the brightness of the image\cite{kim2008adaptive}.
\par
When the image is enhanced to acceptable quality, the classification process is carried out. The different classifications models that are used to predict the classes require feature vectors of respective images. But, because of recent advancement in parallel processing technologies like Graphical Processing Units (GPU) have made the process of training and classification effective and efficient. Google's machine learning API "tensorflow" and standard architecture of GoogleNet \cite{szegedy_liu_jia_sermanet_reed_anguelov_erhan_vanhoucke_rabinovich_2015} has made the application of machine learning much more simpler. The GoogleNet architecture can be tweaked and trained to custom NIR images to achieve higher identification with minimum training \cite{peng_wang_chen_liu_2016}. 
\par
Devices like Field Programmable Gate Arrays (FPGA) are extensively used to develop hardware circuitry that provides optimized environment for real time image enhancement and detection applications\cite{shiehapplying}. However, in this paper we have used a standard Intel laptop for the processing of images and training the classifier. We have implemented some best performing image enhancement algorithms for enhancement of infrared images after the correction of image taken by NIR Pi-Camera using Novel Iterative Method of Calibration \cite{ellmauthaler2013novel}. The enhanced image is then classified using Neural Network Classifier. 
\par
In this paper, the methodology section explains our system and the implementation of algorithms and models that were used. Also, the algorithms and flowcharts are included in this section. The results obtained using different algorithms separately or in complex fashion are then compared using different enhancement metrices and classification accuracy in results and analysis section. Finally, the conclusion section concludes the paper.


\section{Methodology}\label{section:methodology}

\subsection{Image Enhancement}

The tangential and radial distortion produced because of the inaccurate calibration of lenses of camera can be eliminated by correction of the images on the basis of camera parameters obtained after the calibration of IR Camera. Novel Iterative Method of Calibration \cite{ellmauthaler2013novel} of IR Camera determines the intrinsic and extrinsic camera parameters which are based upon pin-hole camera model \cite{elements_of_geometric_computer_vision_2005} by iteratively mapping the 3D world coordinates into 2D camera coordinates. A 2D array of 8\*8 IR-LED bulbs are used as object to determine the coordinates for iteration.
\begin{equation}\label{eq:1}
\mu x = K [ R | t ] X
\end{equation}
\par
 Here, $\mu$ is a scale factor, $x$ is image point (homogeneous 2D vector) in camera coordinate, $K$ is intrinsic camera parameters, $[R | T]$ are rotation | translation coefficients and X is object point vector in world coordinate(3D).
\par 
The image of IR-LED array is first binary thresholded to gray-scale image. The extracted bulb region is then fit into ellipse and the centroid is calculated. Using DLT algorithm \cite{dubrofsky2009homography}, the first homography matrix (H) is obtained which is further refined by minimizing the cost function in equation \ref{eq:2}:
\begin{equation}\label{eq:2}
\sum{||x_i' - H\bar{X}_i||}
\end{equation}
\par 
The final calibration points then calculated using refined \emph{H} and world coordinate which is used to calculate the camera parameters by projecting the images into fronto-parallel plane and iterating until convergence. The obtained parameters are used to correct the images taken by the camera.
\par 
The interpretability and perception of information in image was improved by the means of contrast enhancing algorithms in spatial domain. Some of the enhancement techniques used in this paper were Histogram Equalization \cite{gonzalez2012digital} and its derivatives like Adaptive Histogram Equalization (AHE) \cite{zhu2012adaptive}, Contrast Limited Adaptive Histogram Equalization (CLAHE) \cite{pizer1987adaptive} \cite{reza2004realization}. Entropy and Peak Signal to Noise Ratio(PSNR) were used to demonstrate the enhancement quantitatively.
\par 
Entropy is the information content in any signal which is directly related with randomness of the signal. For images, the more the variation in intensity value across pixels, the more will be the entropy as expressed in equation \ref{eq:3}\cite{kotha2017performance}. Higher entropy, always doesn't mean that the higher information content. It also denotes the higher level of noise in our signal.
\begin{equation}\label{eq:3}
E = \- \sum_{i=0}^{n-1}{p_i\log_{2}{p_i}}
\end{equation} 
Here, $p_i$ is the probability of occurrence of $i^{th}$ gray level.
\begin{equation}\label{eq:4}
p_i = \frac{q_k}{Q}
\end{equation}
\par 
Peak Signal to Noise Ratio between two images is expressed as in equation \ref{eq:5}
\begin{equation}\label{eq:5}
PSNR = 10\log_{10}\left(\frac{{MAX_i}^2}{MSE}\right)
\end{equation}
where $MAX_i$ is the maximum pixel intensity value in image. For 8-bit image, $MAX_i = 2^8 -1 = 255$.
\par 
MSE is the Mean Square Error which is given as in equation \ref{eq:6}
\begin{equation}\label{eq:6}
MSE = \frac{1}{mn} \sum_{i=0}^{m-1} \sum_{j=0}^{n-1}[I(i,j)-K(i,j)]^{2}
\end{equation}
where $I(i,j)$ represents pixel value of original (reference) $mxn$ image at $(i,j)$ position and $K(i,j)$ represents the same for enhanced $mxn$ image at $(i,j)$ position.
\par 
This metric works desirably in cases of measurement of performance of image compression. However, in cases of image enhancement, the measure seems to produce incomprehensible results. So, the metric is modified to fit the need\cite{kotha2017performance}. The modified metric is expressed as in equation \ref{eq:7}
\begin{equation}\label{eq:7}
PSNR_{VAR} = 10\log_{10}\left(\frac{{MAX_i}^2}{var(i)}\right)
\end{equation}
where $var(i)$ represents the variance of pixel values in image.
\par 
This modified metric gives the measure of variance of pixel values from its mean value. 
\par 
Histogram of a image represents the numerical data of intensity and number of pixels corresponding to it. 
\par 
In histogram equalization, the CDF (Cumulative Distribution Function) of the pixel intensity is normalized in such a way that the new CDF becomes the linear function with constant slope. 
\par 
CDF of the image can be determined by using equation in equation \ref{eq:8}
\begin{equation}\label{eq:8}
CDF(i) = \sum_{j=0}^{i}{P_x(j)}
\end{equation}
where,
\begin{equation}\label{eq:9}
P_x(i) = n_i/n
\end{equation}
and $n$ represents the total number of pixels and $n_i$ is the number of pixels having same intensity $i$.
\par 
The goal of Histogram equalization is to recalculate the pixel intensity such that the new CDF is equal to the intensity(i) times any constant value(k).
\begin{equation}\label{eq:10}
CDF_{new}(i) = i*k
\end{equation}
\par 
For that purpose, the pixels intensity has to be normalized as in equation \ref{eq:11}
\begin{equation}\label{eq:11}
I(i) =  \frac{(CDF(i) - CDF_{min})}{(n -1)}  * (2^N -1)
\end{equation}
where $N$ is the bit depth of image.
\par 
In traditional approach to histogram equalization, the entropy of the result image is increased which causes loss of information. The introduction of parameter $\beta$ \cite{zhu2012adaptive} and takes entropy content as target function preserving the entropy described by equation \ref{eq:3}.
\par
$f_i$ is the gray value of $i^{th}$ gray level in the original image. Position $j$ of the resultant image for corresponding $g_j$ of the original image is given by the transformation in equation \ref{eq:12}.
\begin{equation}\label{eq:12}
j = (m-1)\frac{\sum_{k=0}^{i-1}{p_k}}{\sum_{k=0}^{i-1}{p_k} + \sum_{k=i+1}^{m-1}{p_k}}
\end{equation}
\par 
The parameter $\beta$ is introduced to prevent the gray-level with low number of pixels being overwhelmed by gray-level with large number of pixels in the neighborhood. the new transformation becomes equation \ref{eq:13}
\begin{equation}\label{eq:13}
j = (m-1)\frac{\sum_{k=0}^{i-1}{p_k}}{\sum_{k=0}^{i-1}{p_k} + \beta \sum_{k=i+1}^{m-1}{p_k}}
\end{equation}
\par 
Selection of $\beta$ for an 8-bit image with 256 gray-levels can be divided into 3 categories: low gray levels, middle gray level and high gray level. The threshold is set at TL=85, TH=170. The pixels at each of these categories are calculated and recorded. The maximum of the three is found and image type is determined.
\begin{itemize}
	\item $\beta$ = 0.8 if number of pixel in low gray level is highest.
	\item $\beta$ = 1.1 if number of pixel in medium gray level is highest.
	\item $\beta$ = 1.5 if number of pixel in high gray level is highest.
\end{itemize}
\par 
Although the algorithm has been only described for gray-scale image, this idea has been extended and used for all 3 channels\emph{(R,G,B)}\cite{zhu2012adaptive}.
\par
Conventional histogram equalization operates by aiming to equalize the histogram of entire image \emph{i.e} global contrast of the image is considered. This technique yields good result if the contrast of the image is uniform or nearly uniform over the entirety of the image. However, if the image has regions of sharp difference in contrast, then this technique will usually result in blowing up of detail in bright areas. Contrast Limited Adaptive Histogram Equalization (CLAHE) was developed to address this issue\cite{pizer1987adaptive} \cite{reza2004realization}.
\par
Starting with selection the grid size(minimum $32$ X $32$) depending upon the dimension of image, grid points are identified from the top-left corner in CLAHE. Histogram for each grid point is calculated separately and clipped if above the certain level. The new histogram is used to calculate CDF for each pixel by making it a grid point. The neighboring 4 grid points are taken for each pixel and interpolation of intensity through CDF mapping gives the final pixel intensity.

\subsection{Object Recognition}
\par
Convolutional Neural Network (CNN) is an advanced form of feedforward artificial neural network consisting of convolutional layers capped by fully connected layers which are useful for features with local relationship and thus a common choice for image classification\cite{haykin2004comprehensive}.

\par
The CNNs exploit Local Connectivity and Parameter Sharing scheme. Local Connectivity describes how a set of 2D weights called as filters or kernels connect only to a small region of input called “receptive field” and these filters are subsequently convolved over the entire image. Parameter Sharing scheme describes how the weights of the filter set can be shared among the filters of same channel or in other words slide the filter over the image which greatly reduces the number of parameters.
\par
ConvNets are basically composed of initial convolutional layers which serve as feature extractor and final fully connected layers which serve as classifiers. The convolutional layers are often present in conjunction with activation layers and pooling layers which then produce features with reduced dimension. These set of layers are often repeated. 
\par
An intuitive understanding of the features produced by the individual set of layers could be attained by assuming the earlier filters to detect simple features like lines curves, and latter filters to to detect more complex features like shapes which is expected with the increased depth of filter. Conventional idea for a ConvNet is to use a fully-connected layer (typical neural net) to terminate the ConvNet model which perform actual classification. This layer is fed with the down-sampled features from previous convolutional layers serving to extract feature from images.
\par
As mentioned earlier, a CNN consists of convolutional layers and fully connected layers contributing to a big number of weights and biases. A dataset sufficient for training a CNN with random initialization of weights and biases is rare. Moreover, it is computationally prohibiting for training complex a CNN within limited time.
\par 
This arduous task of finding a suitable dataset and computation resources is greatly minimized by the advent of Transfer Learning\cite{cs231n_cnn}. It involves tweaking of a pre-trained model with limited user dataset. Transfer Learning is implemented in terms of using CNN as fixed feature extractor and fine tuning CNN.
\par 
By removing the final fully connected layer of a pre-trained CNN, the rest of the network can be used as a fixed feature extractor for new dataset. This tweaked network produces features from the dataset also known as CNN Codes. The obtained CNN codes can then be fed to a fully connected network with required number of classes\cite{cs231n_cnn}.
\par 
Instead of random initialization of weights and biases, the model can be initialized using the weights and biases of the pre-trained network. The parameters are then fine tuned using the new dataset. Parameters of initial layers can be kept constant to prevent over-fitting. It is observed that initial convolutional layers detect generic features from images and high level layers detect more specific features. This method however, requires bigger dataset and is more computationally intensive\cite{cs231n_cnn}.
\par
The implementation of 2015 iteration of GoogLeNet, Inception-V3 Model\cite{szegedy2016rethinking} was used in our project with minor tweaking. The final fully connected classifier layer was detached from the model and our own fully connected layer with our appended to the CNN layer. The training of the model was carried out for our custom dataset.
\par
The inception-V3 structure consists of multiple layers cascaded together. The basic building block of inception v3 is the inception module. The inception module is combination of multiple operations done. They typically involve 1x1 convolutions in parallel with two more 1x1 convolutions and a 3x3 max pooling. The outputs of the latter 2 1x1 convolutions are individually passed to other convolutions. 
\par
In our implementation, the conv-net produces a feature matrix of 2048 elements whose flattened form is input for the final fully connected layer. The layer of dimension 2048x5 is thus trained to be optimized for giving the output among 5 classes by 500 images of each class. The output of this layer is a vector with prediction for the 5 classes which is fed to softmax layer to give relative prediction (which sums up to a 100\%) for the classes.

\subsection{Our System}

\begin{figure}[ht]
\centering
\scalebox{0.75}{
\begin{tikzpicture}[block/.style={draw, thick, minimum height=1cm, align=center},start/.style={draw, thick, rounded corners, minimum width=3cm, minimum height=1cm,thick,align= center}]
	\node[start](start){Start};
	\node[block, below= of start](capture){Capture the IR image};
	\node[block, below= of capture](calib){Calibrate the IR image};
	\node[block, below= of calib](enhance){Enhance the IR image};
	\node[block, below= of enhance](cnn){Convolutional Neural Network (Feature Extraction)};
	\node[block, below= of cnn](classify){Fully Connected Layer (Classification)};
	\node[start, right= 1.5cm of classify](end){End};

	\draw[->] (start) -- (capture);
	\draw[->] (capture) -- (calib);
	\draw[->] (calib) -- (enhance);
	\draw[->] (enhance) -- (cnn);
	\draw[->] (cnn) -- (classify);
	\draw[->] (classify) -- (end);

	\node[draw,dashed,inner xsep=2mm,inner ysep=2mm,fit=(calib)(enhance)](dotenhance){};
  	\node[draw,dashed,inner xsep=2mm,inner ysep=2mm,fit=(cnn)(classify)](dotclassify){};
  	\node at (dotenhance.east) [right, inner sep=1mm] {Enhancement};
	\node at (dotclassify.east) [right, inner sep=1mm] {Classification};
  	

\end{tikzpicture}}
\caption{System Block Diagram}
\label{fig:3}
\end{figure}
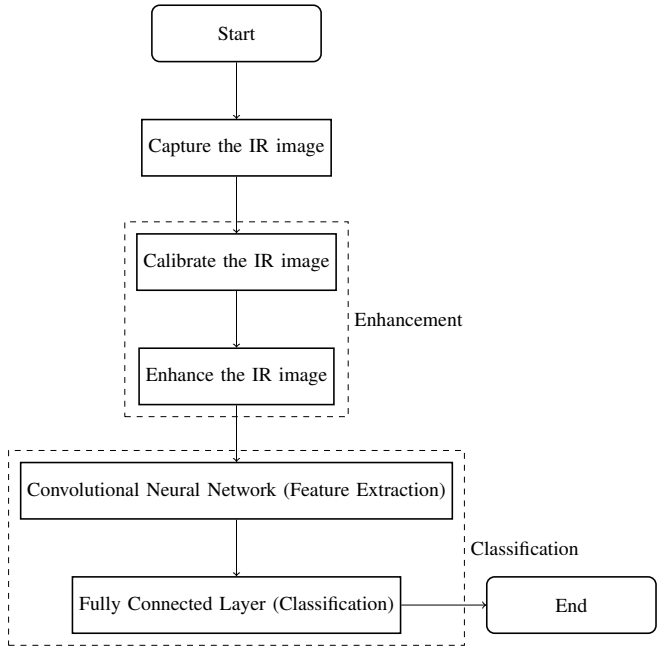

\par
Our setup for this study involved a completely light-sealed box of (0.6mx0.6mx0.3m) dimension wherein IR LEDs were dimly lit internally for passive illumination. This setup was designed to replicate the situation of surveillance in low light situation where the pictures of subject were taken and sent to remote computer for enhancement and classification. Our system block diagram can be seen in figure \ref{fig:3}.


\section{Results and Analysis}
\par
The primary problems that can be handled by software camera calibration technique are: tangential and radial distortion. The parameters $c_z$ and $c_y$ represents the distortion center or the principle point. In our case, the principal point was found to be (501.48,378.45). 
\par
\begin{figure}[ht]
	\centering
  \begin{subfigure}[b]{0.2\textwidth}
    \includegraphics[width=\textwidth]{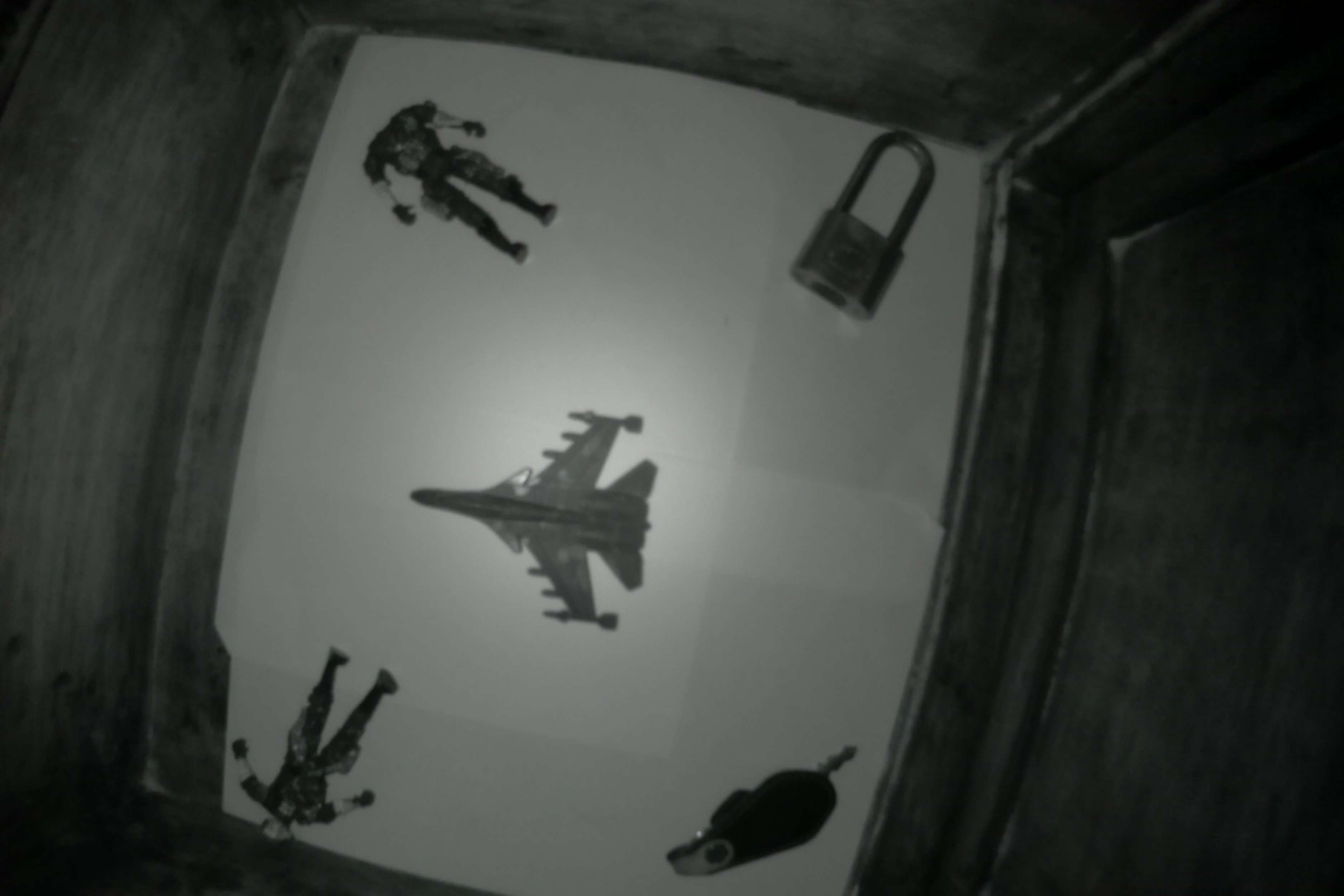}
    \caption{Uncalibrated Image}
    \label{fig:4a}
  \end{subfigure}
  \hfill
  \begin{subfigure}[b]{0.2\textwidth}
    \includegraphics[width=\textwidth]{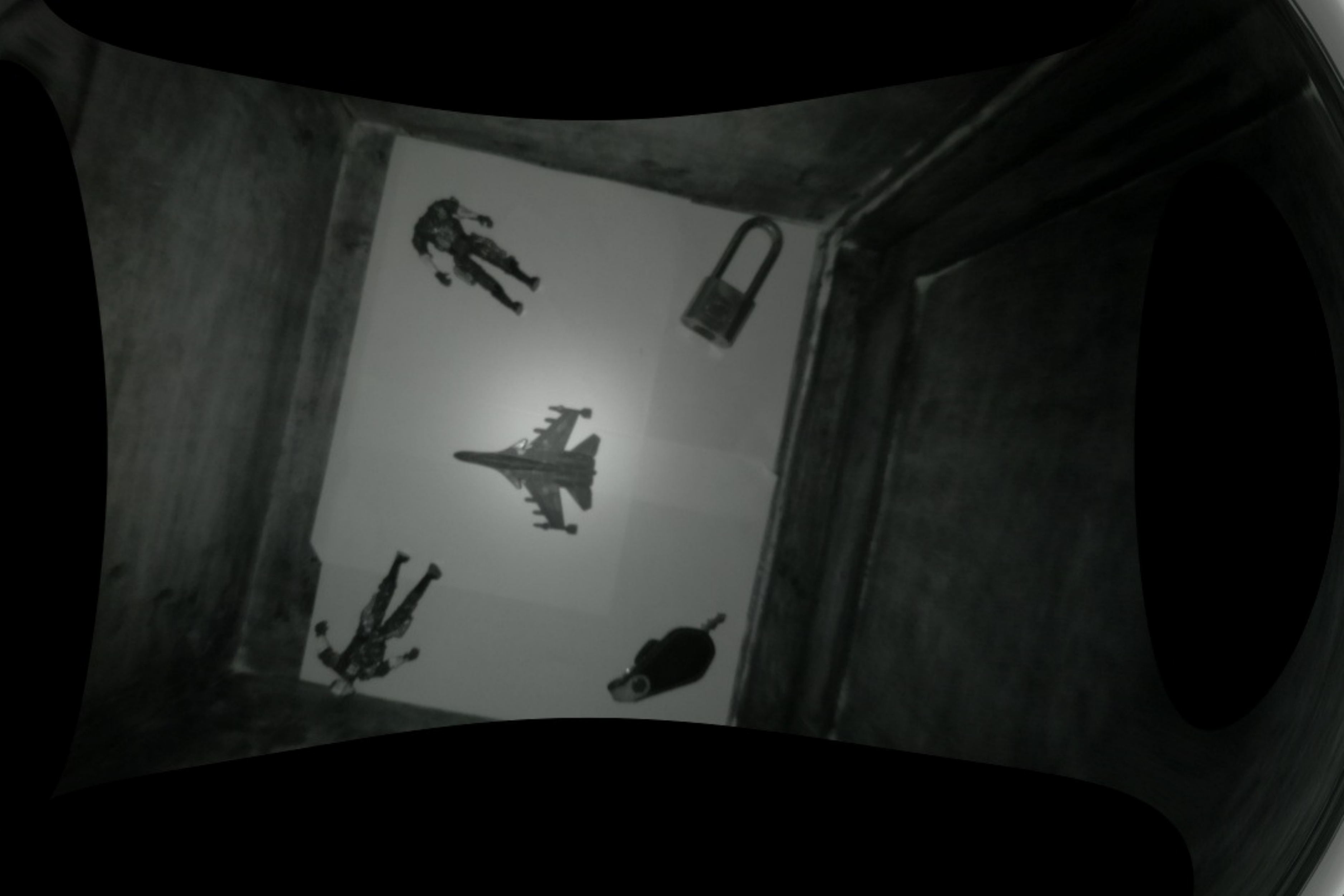}
    \caption{Calibrated Image}
    \label{fig:4b}
  \end{subfigure}

  \begin{subfigure}[b]{0.2\textwidth}
    \includegraphics[width=\textwidth]{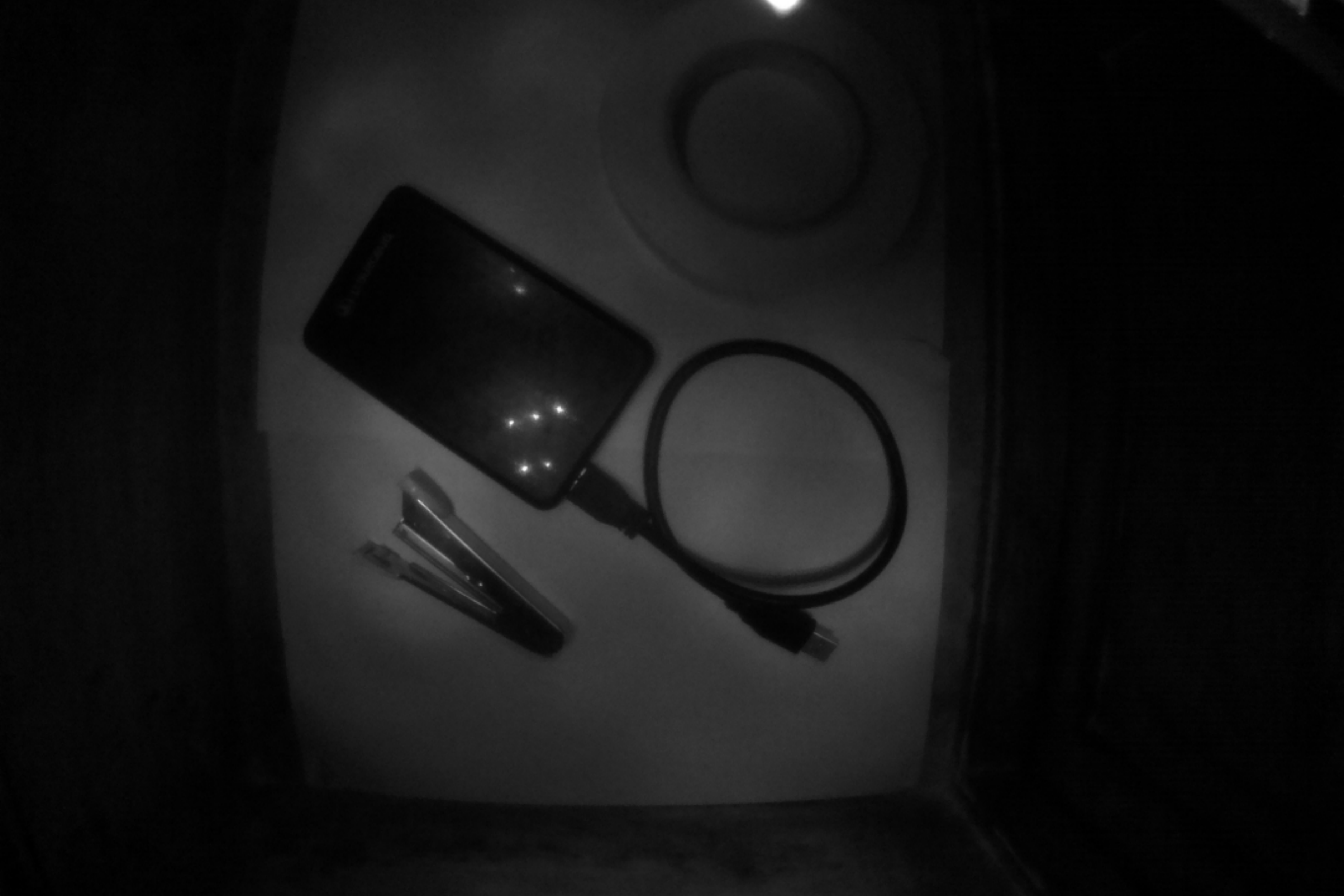}
    \caption{Uncalibrated Image}
    \label{fig:4c}
  \end{subfigure}
  \hfill
  \begin{subfigure}[b]{0.2\textwidth}
    \includegraphics[width=\textwidth]{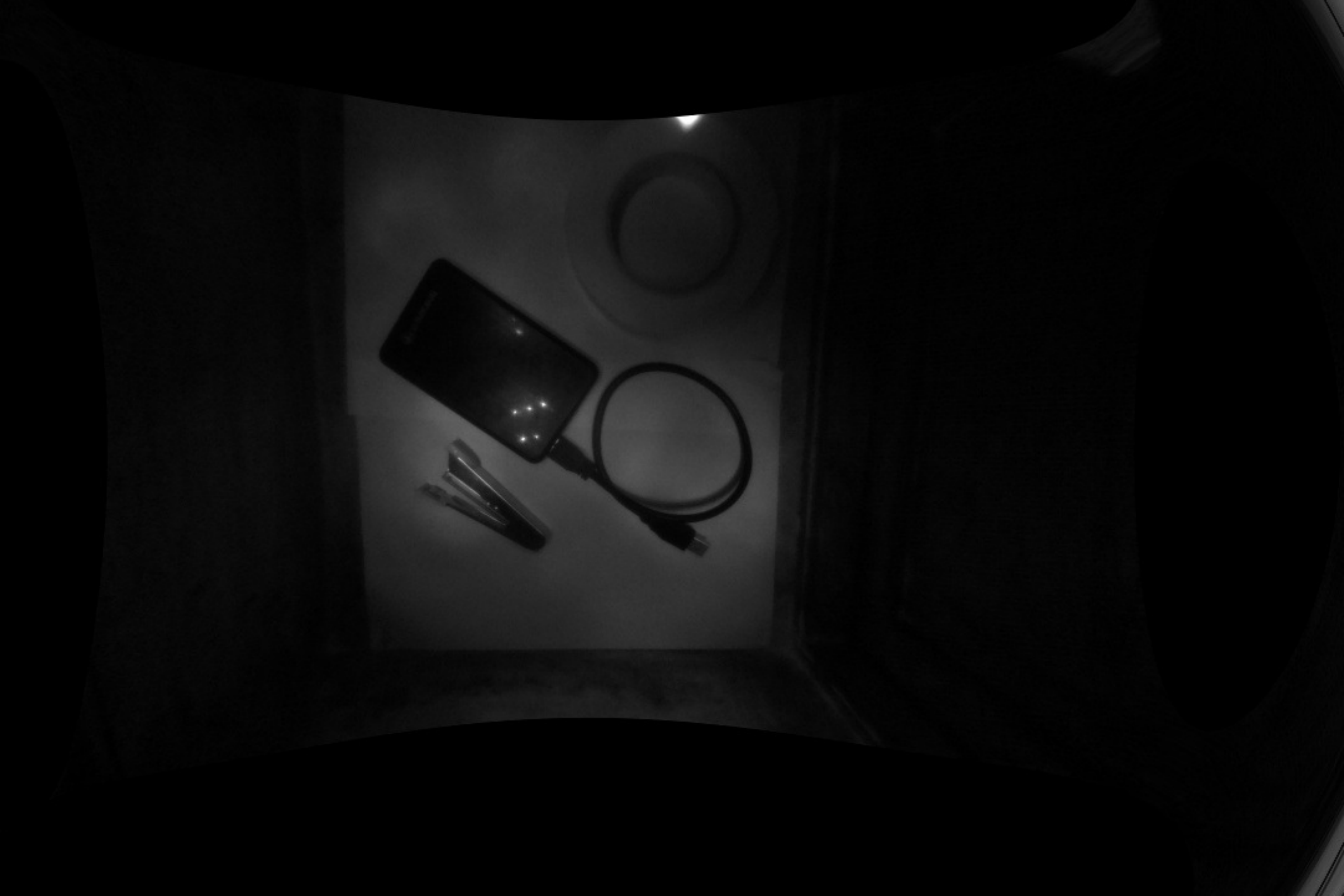}
    \caption{Calibrated Image}
    \label{fig:4d}
  \end{subfigure}

  \begin{subfigure}[b]{0.2\textwidth}
    \includegraphics[width=\textwidth]{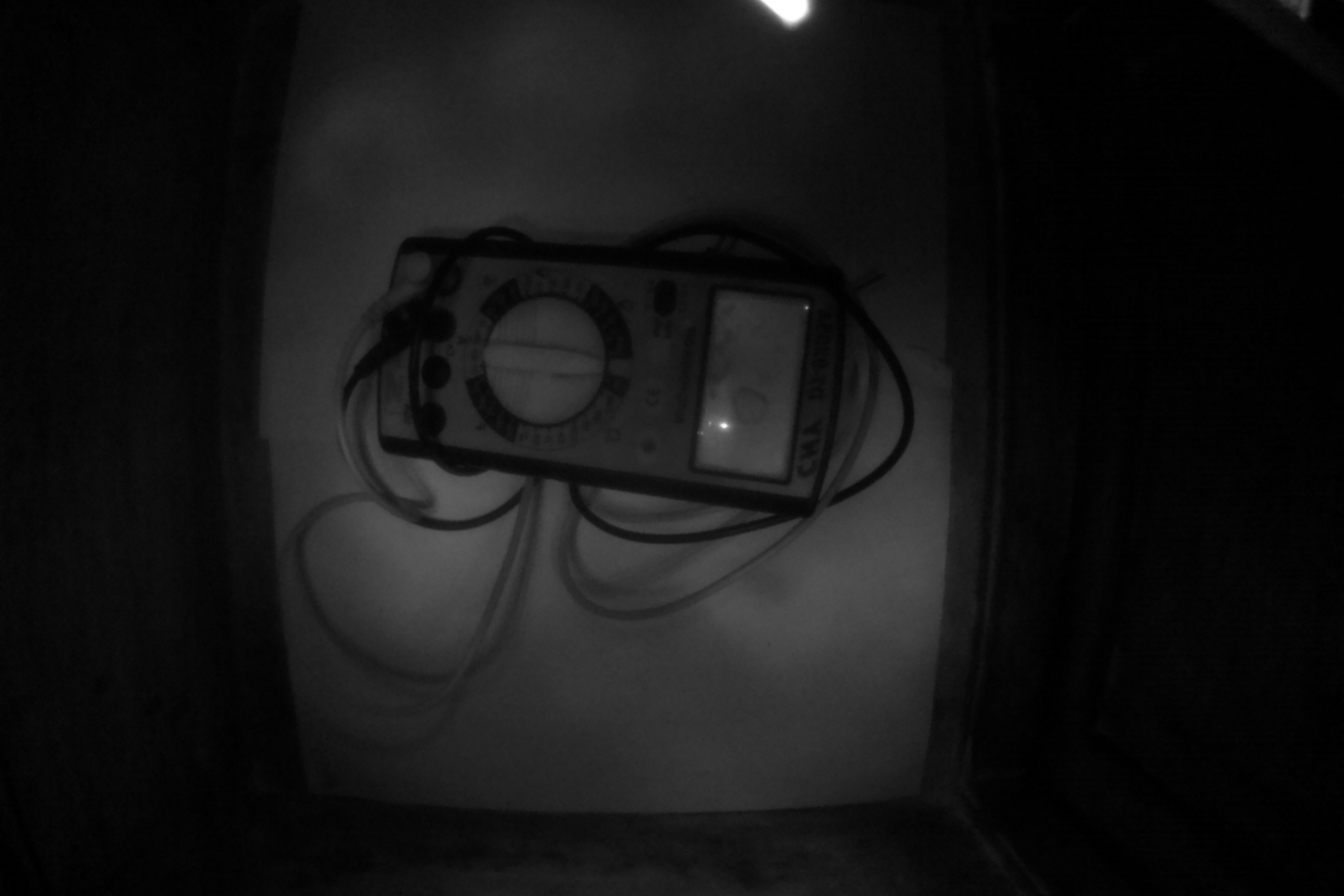}
    \caption{Uncalibrated Image}
    \label{fig:4e}
  \end{subfigure}
  \hfill
  \begin{subfigure}[b]{0.2\textwidth}
    \includegraphics[width=\textwidth]{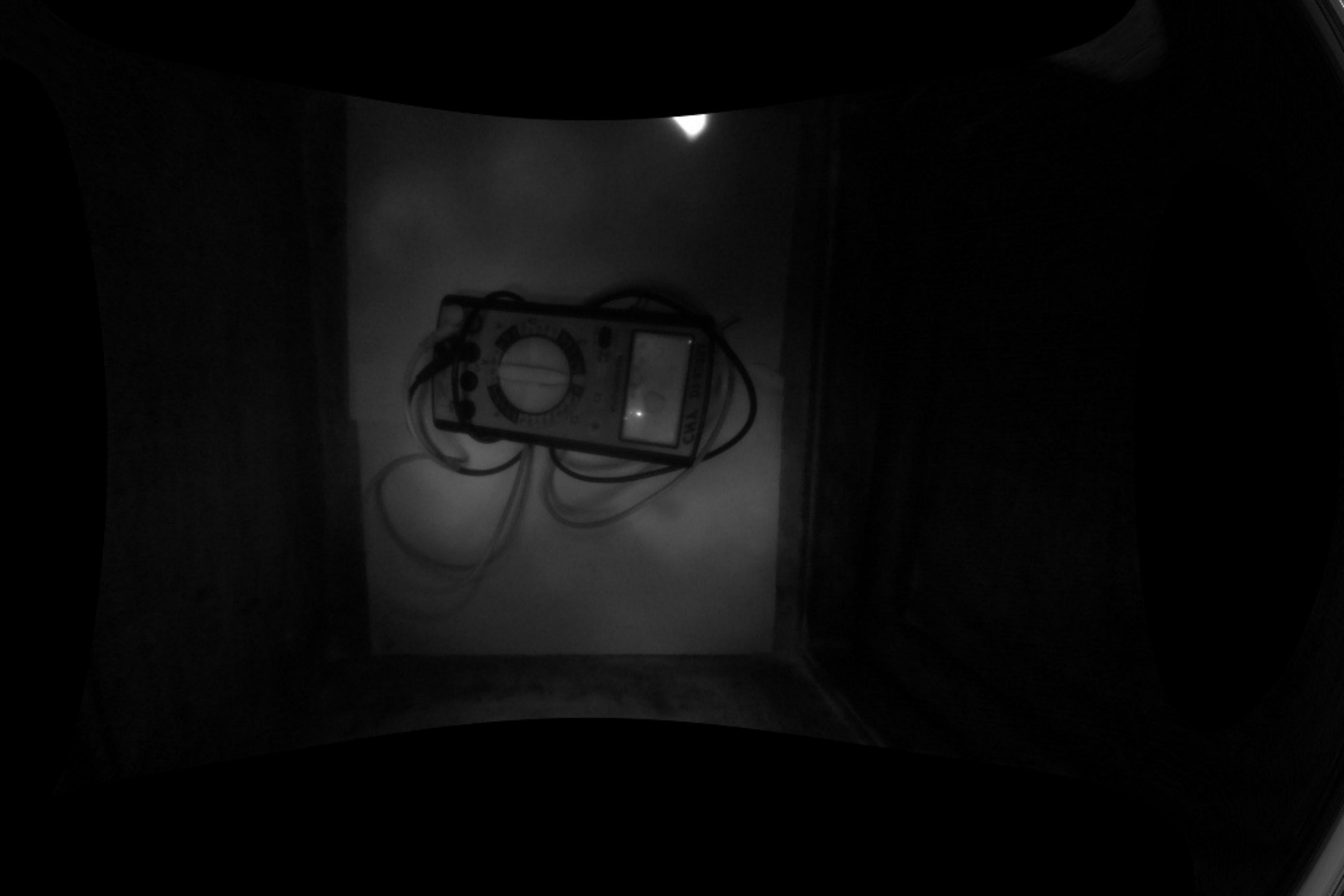}
    \caption{Calibrated Image}
    \label{fig:4f}
  \end{subfigure}  
  \caption{Images before and after calibration}
  \label{fig:4}
\end{figure}

The cause of tangential distortion can be traced to the misalignment of the lens and the sensor-plane. The values obtained for tangential distortion coefficients were $p_1$ and $p_2$ as 0.00174822 and 0.00352084. This results shows that the camera produces very small tangential distortion. However, the radial distortion is significant in the camera unit that we are using. The radial distortion coefficients are $k_1$, $k_2$ and $k_3$ with values of respectively -0.3439249, 0.1697238 and -0.0360944 respectively. Radial distortion produces bulging out effect in the images that we have taken. This is evident in figure \ref{fig:4}.
\begin{table}[H]
  \centering
  \caption{Intrinsic parameters of IR camera}
  \label{tab:intrinsic table}
  \begin{tabular}{|c|c|c|}
    \toprule
    Parameter & value (Pixel Units)\\
    \midrule
    focal length of lens along x-axis ($f_x$) & 612.383958\\
    focal length of lens along y-axis ($f_y$) & 611.2666744\\
    principle point along x-axis ($c_z$) & 501.484677\\
    principle point along y-axis ($c_y$) & 378.459481\\
    \bottomrule
  \end{tabular}
\end{table}

\begin{table}[H]
  \centering
  \caption{Extrinsic parameters of IR camera}
  \label{tab:extrinsic table}
  \begin{tabular}{|c|c|c|}
    \toprule
    Parameter & value(Pixel units)\\
    \midrule
    radial distortion parameter ($k_1$) & -0.3439249\\
    radial distortion parameter ($k_2$) & 0.1697238\\
    radial distortion parameter ($k_3$) & -0.0360944\\
    tangential distortion parameter ($p_1$) & 0.00174822\\
    tangential distortion parameter ($p_2$) & 0.00352084\\
    \bottomrule
  \end{tabular}
\end{table}

\par
These effects have been ameliorated by using the radial and tangential distortion parameters to undistort the image. These parameters are used to correct the bulging effect. The straight lines seems to appear straight in the corrected image.
\par
The quantitative parameters like entropy of the image, Mean Square Error(MSE), Peak Signal to Noise Ratio (PSNR) and modified PSNR with variance of the image are chosen to study the effect of image enhancement parameters.
\par
\begin{figure}[ht]
\centering
\begin{subfigure}[b]{0.2\textwidth}
\includegraphics[width=\textwidth]{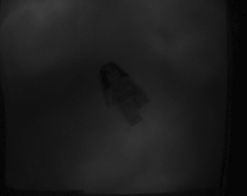}
\caption{Original Image}
\label{fig:5a}
\end{subfigure}
\hfill
\begin{subfigure}[b]{0.2\textwidth}
\includegraphics[width=\textwidth]{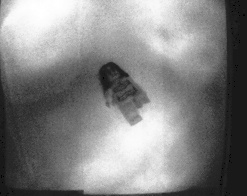}
\caption{Image after Histogram Equalization}
\label{fig:5b}
\end{subfigure}

\begin{subfigure}[b]{0.2\textwidth}
\includegraphics[width=\textwidth]{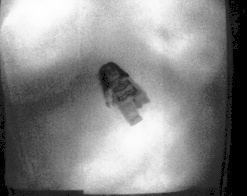}
\caption{Image after Adaptive HE}
\label{fig:5c}
\end{subfigure}
\hfill
\begin{subfigure}[b]{0.2\textwidth}
\includegraphics[width=\textwidth]{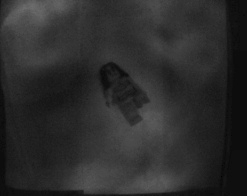}
\caption{Image after CLAHE}
\label{fig:5d}
\end{subfigure}
\caption{Image after different enhancement techniques}
\label{fig:5}
\end{figure}

Entropy is simply the measure of randomness or information content in the image. In image processing domain, entropy gives the measure of how the pixel values differ from one another. A highly homogeneous image like completely dark or completely bright image carries no information and therefore entropy value is very low. This can be seen for original image of figure \ref{fig:5a}, in table \ref{tab:enhance} that it has the least entropy of 1.06021 due to the lowest brightness. These images are taken in low lighting condition and are predominantly dark i.e high number of pixels tend to be in low intensity region. Increase in entropy can improve the visual appeal and increase the information in the image. This can be seen by images enhanced using HE, AHE and CLAHE. The entropy of these images being 1.79100, 1.7581 and 1.45936 respectively. However, high entropy doesn't always guarantee more pleasing images as high entropy is also caused due to presence of noise as seen in figure \ref{fig:5b}(AHE). So, Adaptive Histogram Equalization which is based on increasing the image entropy might not always provide with desirable outcomes in all scenarios.
\par
MSE and PSNR is also used in conjunction with entropy to get a better idea of the results of enhancement. MSE is one of basic measure of error between the new image and the standard image. Lower value of MSE is better. We have taken the images of objects by properly illuminating them i.e. in good lighting condition. The poorly lit images are enhanced and MSE is computed with the properly lit images. In the scenarios we used, there were regions of very high intensity and very dark intensity. In this constraint, CLAHE has outperformed other enhancement techniques which can be seen in table \ref{tab:enhance}. MSE of images obtained using CLAHE was 264 which is significantly lower than its counterparts. This results can also be expressed in terms of PSNR in which the MSE is seen in denominator term. So, higher PSNR is better. The results obtained using MSE is exactly reflected while using PSNR. CLAHE has yielded higher PSNR in cases of non-uniform illumination with values of 23.9154.

\begin{table}[H]
	  \centering
	  \caption{Parameter chart for a sample image}
	  \label{tab:enhance}
	  \begin{tabular}{|c|c|c|c|c|}
	    \toprule
	    Enhancement & entropy & MSE & PSNR(dB) & PSNR-VAR(dB)\\
	    \midrule
	    Original & 1.06021 & 771 & 19.259 & 32.4014\\
	    HE & 1.79100 & 13713 & 6.7592 & 11.0606\\
	    AHE & 1.7581 & 14177 & 6.6146 & 9.7598\\
	    CLAHE & 1.45936 & 264 & 23.9154 & 25.0872\\
	    \bottomrule
	  \end{tabular}
\end{table}

\par
Another measure using modified PSNR is used. This parameter is used to measure the variance of the image pixels. Higher variance in the image pixels suggests the contrast is well-stretched and lower variance in the image suggests opposite. Since, the variance term is introduced in the denominator, the results has to be inferred accordingly. The algorithm yielding closer PSNR-VAR value to that of original properly lit image is considered better in which CLAHE has outperformed other algorithms with value of 25.0872 which is close to 20.47 of the properly lit image.

After enhancement, the images were sent to classifier. The softmax function of the fully connected layer gave the prediction of the class. For the images obtained after enhancement, the classifier gave the results as in table \ref{tab:accuracy}. The classification accuracies were 58\%, 71\%, 74\% and 85\% for no enhancement, histogram equalization, adaptive histogram equalization and contrast adaptive histogram equalization respectively. From the results, it can be said that CLAHE being superior in enhancement clearly outperforms other enhancement techniques for the purpose of enhancing IR images and as preprocessing step for classification for our case.

\begin{table}[H]
	  \centering
	  \caption{Classification}
	  \label{tab:accuracy}
	  \begin{tabular}{|c|c|}
	    \toprule
	    Enhancement & Accuracy \\
	    \midrule
	    Original & 58\% \\
	    HE & 71\% \\
	    AHE & 74\% \\
	    CLAHE  & 85\% \\
	    \bottomrule
	  \end{tabular}
	\end{table}


\section{Conclusion}
\par
 We designed a system capable of mimicking low light scene and capturing images at the varying level of illumination. These images were subjected to different image enhancement algorithms that worked on different principles like stretching the contrast of the original image to make it more palatable, increasing the entropy of the original image to boost the information content or add randomness. The results are quite exciting in this part.
\par
 We implemented a Convolutional Neural Network capable of detecting objects present in the scene and predicting class label to these objects. We implemented transfer learning to devise a CNN with limited dataset and ordinary processing unit (4 GB RAM and no graphics card). In this way, we demonstrated a proof of concept that systems capable of enhancing and detecting the objects in low light scenarios. The  accuracy of the classifications were 58\%, 71\%, 74\% and 85\% for no enhancement, histogram equalization, adaptive histogram equalization and contrast adaptive histogram equalization respectively. Therefore, image enhancement techniques can be employed to improve the classification accuracy produced by a specialized CNN devised using transfer learning technique.

 \par
 There lies tremendous possibility in the field of night vision surveillance and we could only scratch the surface. We are determined to continue our effort in the field of night vision surveillance and improve upon the existing technologies and concepts in this field.  


\section{Acknowledgments}
\par
This paper is an excerpt of final year project of Bachelor in Electronics and Communication Engineering, submitted to the Department of Electronics and Computer Engineering, TU, IOE, Pulchowk Campus in the year 2017. This work was supervised by Mr. Dinesh Baniya Kshatri. The authors would like to express their sincerest gratitude the department of Electronics and Computer Engineering, IoE, Pulchowk Campus.

\nocite{*}
\bibliographystyle{./bibliography/IEEEtran}
\bibliography{./bibliography/references}


\end{document}